\documentclass[11pt,a4paper]{article}
\usepackage[hyperref]{emnlp-ijcnlp-2019}
\pdfoutput=1
\usepackage{times}
\usepackage{latexsym}
\usepackage{amsmath}
\usepackage{url}
\usepackage{multirow}
\usepackage{xr}
\usepackage{array}
\usepackage{pgfplots}
\usepackage{amsfonts}

\newcolumntype{C}[1]{>{\centering\arraybackslash}p{#1}}

\aclfinalcopy 

\title{Automatic Argument Quality Assessment - New Datasets and Methods}

\author {
 Assaf Toledo\thanks{\ \ These authors equally contributed to this work.}\hspace{0.15cm},\hspace{0.2cm}Shai Gretz$^{*}$, Edo Cohen-Karlik$^{*}$, Roni Friedman$^{*}$, Elad Venezian,\\
 \textbf{Dan Lahav, Michal Jacovi, Ranit Aharonov and Noam Slonim}\\
 IBM Research
}

\date{}

\begin{document}
\maketitle

\begin{abstract}
We explore the task of automatic assessment of argument quality. To that end, we actively collected $6.3k$ arguments, more than a factor of five compared to previously examined data. 
Each argument was explicitly and carefully annotated for its quality. In addition, $14k$ pairs of arguments were annotated independently, identifying the higher quality argument in each pair. In spite of the inherent subjective nature of the task, both annotation schemes led to surprisingly consistent results. We release the labeled datasets to the community. Furthermore, we suggest neural methods based on a recently released language model, for argument ranking as well as for argument-pair classification. In the former task, our results are comparable to state-of-the-art; in the latter task our results significantly outperform earlier methods.
\end{abstract}
\section{Introduction}

Computational argumentation has been receiving growing interest in the NLP community in recent years \cite{ACL-16-tutorial}. With this field rapidly expanding, various methods have been developed for sub-tasks such as argument detection \cite{LippiTorroni,Levy14,rinott15}, stance detection \cite{bar-haim17} and argument clustering \cite{GurevychClustering19}.

Recently, IBM introduced \textit{Project Debater}, the first AI system able to debate humans on complex topics. The system participated in a live debate against a world champion debater, and was able to mine arguments, use them for composing a speech supporting its side of the debate, and also rebut its human competitor.\footnote{For more details: \url{https://www.research.ibm.com/artificial-intelligence/project-debater/live/}} The underlying technology is intended to enhance decision-making.

More recently, IBM also introduced \textit{Speech by Crowd}, a service which supports the collection of free-text arguments from large audiences on debatable topics to generate meaningful narratives. 
A real-world use-case of Speech by Crowd is 
in the field of civic engagement, where the aim is to exploit the wisdom of the crowd to enhance decision making on various topics. There are already several public organizations and commercial companies in this domain, e.g., Decide Madrid\footnote{\url{https://decide.madrid.es}} and Zencity.\footnote{\url{https://zencity.io}} As part of the development of Speech by Crowd, $6.3k$ arguments were collected 
from contributors of various levels, 
and are 
released as part of this work.

An important sub-task of such a service is the automatic assessment of argument quality, which has already shown its importance for prospective applications such as automated decision making \cite{BenchCapon09}, argument search \cite{WachsmuthWeb}, and writing support \cite{StabAndGur}. Identifying argument quality in the context of Speech by Crowd allows for the top-quality arguments to surface out of many contributions. 


Assessing argument quality 
has driven practitioners in a plethora of fields for centuries – from philosophers \cite{Aristotle07}, through academic debaters, to argumentation scholars \cite{Walton2008}. An inherent difficulty in this domain is the presumably subjective nature of the task. \newcite{Wachsmuth2017} proposed a taxonomy of quantifiable dimensions of argument quality, comprised of high-level dimensions such as \textit{cogency} and \textit{effectiveness}, and sub-dimensions such as  \textit{relevance} and \textit{clarity}, that together enable the assignment of a holistic quality score to an argument.

\newcite{Gurevych16} and \newcite{Gurevych18} take a relative approach and treat the problem as relation classification. They focus on \textit{convincingness} -- a primary dimension of quality -- and determine it by comparing pairs of arguments with similar stance. In this view, the convincingness of an individual argument is a derivative of its relative convincingness: arguments that are judged as more convincing 
when compared to others are attributed higher scores. These works explore the labeling and automatic assessment of argument convincingness using two datasets introduced by \newcite{Gurevych16}: \textit{UKPConvArgRank} (henceforth, \textit{UKPRank}) and  \textit{UKPConvArgAll}, which contain $1k$ and $16k$ arguments and argument-pairs, respectively.


\newcite{gleize19} also take a relative approach to argument quality, focusing on ranking convincingness of \textit{evidence}. Their solution is based on a Siamese neural network, which outperforms the results achieved in \newcite{Gurevych18} on the \textit{UKP} datasets, as well as several baselines on their own dataset, \textit{IBM-ConvEnv}.\footnote{As this work is relatively recent and was published after our submission, we were not able to compare to it.}

Here, we extend earlier work in several ways: (1) introducing a large dataset of actively collected arguments, carefully annotated for quality; 
(2) suggesting a method for argument-pair classification, which outperforms state-of-the-art accuracy on available datasets;
(3) suggesting a method for individual argument ranking, which achieves results comparable to the state of the art. 

Our data was collected actively, via a dedicated user interface. This is in contrast to previous datasets, which were sampled from online debate portals. We believe that our approach to data collection is more controlled and reduces noise in the data, thus making it easier to utilize it in the context of learning algorithms (see Section \ref{sec:data-comp}). 

Moreover, we applied various cleansing methods to ensure the high quality of the contributed data and the annotations, as detailed in Section \ref{dataCleansing}.

We packaged our data in the following  datasets, which are released to the
research 
community\footnote{\url{https://www.research.ibm.com/haifa/dept/vst/debating\_data.shtml\#Argument Quality}\label{footdown}}:

\begin{itemize}
    \item \textit{IBM-ArgQ-6.3kArgs} - the full dataset, comprised of all $6.3k$ arguments that were collected and annotated with an individual quality score in the range $[0,1]$.
    \item \textit{IBM-ArgQ-14kPairs} - $14k$ argument pairs annotated with a relative quality label, indicating which argument is of higher quality.
    \item \textit{IBM-ArgQ-5.3kArgs} - the subset of $5.3k$ arguments from \textit{IBM-ArgQ-6.3kArgs} that passed our cleansing process. This set is used in the argument ranking experiments in Section \ref{exp:ranking}, henceforth: \textit{IBMRank}.
    \item \textit{IBM-ArgQ-9.1kPairs} - the subset of $9.1k$ argument pairs 
    from \textit{IBM-ArgQ-14kPairs} that passed our cleansing process, used in the argument-pair classification experiments in Section \ref{sec:argpair}. Henceforth: \textit{IBMPairs}.
\end{itemize}


The dataset \textit{IBMRank} differs from \textit{UKPRank} in a number of ways. Firstly, \textit{IBMRank} includes $5.3k$ arguments, which make it more than 5 times larger than \textit{UKPRank}. Secondly, the arguments in \textit{IBMRank} were collected actively from contributors. Thirdly, \textit{IBMRank} includes {\it explicit\/} quality-labeling of all individual arguments, which is absent from earlier data, enabling us to explore the potential of training quality-prediction methods on top of such labels, presumably easier to collect.

Finally, with the abundance of technologies such as automated personal assistants, we envision automated argument quality assessment expanding to applications that include oral communication. 
Such use-cases pose new challenges, overlooked by prior work, 
that mainly focused on written arguments. As an initial attempt to address these issues, in the newly contributed data we guided annotators to assess the quality of an argument within the context of using the argument as-is to generate a persuasive {\it speech\/} on the topic. Correspondingly, we expect 
these 
data to reflect additional quality dimensions -- e.g., a quality premium on efficiently phrased arguments, and low tolerance to blunt mistakes such as typos that may lead to poorly stated arguments.

\section{Argument Collection}
\label{Data}


As part of the development of Speech by Crowd, online and on-site experiments have been conducted, enabling to test the ability of the service to generate a narrative based on collected arguments. Arguments were collected from two main sources: (1) debate club members, including all levels, from novices to experts; and 
(2) a broad audience of people attending the experiments.   

For the purpose of collecting arguments, we first selected $11$ well known controversial concepts, 
common in the debate world, such as \textit{Social Media}, \textit{Doping in Sports} and \textit{Flu vaccination}. Using debate jargon, each concept 
is used to phrase two ``motions", 
by proposing two specific and opposing policies or views 
towards that concept. 
For example,
for the concept 
\textit{Autonomous Cars}, we suggested 
the motions \textit{We should promote Autonomous Cars} and \textit{We should limit Autonomous Cars}.\footnote{\newcite{Gurevych16} uses the term \textit{topic} for what we refer to as \textit{motion}.} The full list of motions appears in Table \ref{table:args_count} with the number of arguments collected for each.\footnote{In Table \ref{table:args_count}, \textit{vvg} stands for \textit{violent video games.}}



\textbf{Guidelines}
Contributors were invited to a dedicated user interface 
in which they were guided to contribute arguments per concept, 
using the following concise instructions:
\\
\noindent
\setlength{\fboxrule}{1.5pt}
\fbox{
  \parbox{18.6em}{
    You can submit as many arguments as you like, both pro and con, using original language and no personal information (i.e. information about an identifiable person).
  }
}
\\
\\
\noindent
In addition, to exemplify the type of arguments that we expect to receive, contributors were shown an example of one argument related to the 
motion,
provided by a professional debater. The arguments collected had to have $8$ -- $36$ words, aimed at obtaining efficiently phrased arguments 
(longer/shorter arguments were rejected by the UI).
In total, we collected $6,257$ arguments.  

\begin{table}[!htb]
\centering
\small
\begin{tabular}{|p{6cm}|c|} 
\hline
\textbf{Motion} & \textbf{\#Args} \\
\hline
Flu vaccination should be mandatory & 204 \\
\hline
Flu vaccination should not be mandatory & 174 \\
\hline
Gambling should be banned & 342 \\
\hline
Gambling should not be banned & 382 \\
\hline
Online shopping brings more harm than good & 198 \\
\hline
Online shopping brings more good than harm & 215 \\
\hline
Social media brings more harm than good & 879 \\
\hline
Social media brings more good than harm & 686 \\
\hline
We should adopt cryptocurrency & 172 \\
\hline
We should abandon cryptocurrency & 160 \\
\hline
We should adopt vegetarianism & 221 \\
\hline
We should abandon vegetarianism & 179 \\
\hline
We should ban the sale of vvg to minors & 275 \\
\hline
We should allow the sale of vvg to minors & 240 \\
\hline
We should ban fossil fuels & 146 \\
\hline
We should not ban fossil fuels & 116 \\
\hline
We should legalize doping in sport & 212 \\
\hline
We should ban doping in sport & 215 \\
\hline
We should limit autonomous cars & 313 \\
\hline
We should promote autonomous cars & 480 \\
\hline
We should support information privacy laws & 355 \\
\hline
We should discourage information privacy laws & 93 \\
\hline
\end{tabular}
\caption{Motion list and statistics on data collection.}
\label{table:args_count}
\end{table}

\section{Argument Quality Labeling} 
\label{dataCleansing}
We explored two approaches to labeling argument quality: (a) labeling individual arguments (absolute approach): each individual argument is 
directly labeled for its quality;
and (b) labeling argument pairs (relative approach): each argument pair is labeled for which of the two arguments is of higher quality. In this section we describe the pros and cons of each 
approach as well as the 
associated labeling process. 

\subsubsection*{Approaches to Argument Quality Labeling}
The effort in labeling individual arguments scales linearly with the number of arguments, compared to the quadratic scaling of labeling pairs (within the same motion); thus, it is clearly more feasible when considering a large number of arguments. However, the task of determining the quality of arguments in isolation 
is 
presumably 
more challenging; it requires to evaluate the quality of an argument without a clear reference point (except for the motion text). This is where the relative approach has its strength, as it frames the labeling task in a specific context of two competing arguments, and is expected to yield higher inter-annotator agreement. 
Indeed, a comparative approach is widely used in many NLP applications, e.g. in \newcite{Chen:2013} for assessing reading difficulty of documents and in \newcite{Aranberri2017} for machine translation.
 In light of these considerations, here we decided to investigate and compare both approaches. We used the Figure Eight platform\footnote{\url{http://figure-eight.com/}}, with a relatively large number of $15-17$ annotators per instance, 
 to improve the reliability of the collected annotations. 

\subsection{Labeling Individual Arguments}
\label{sec:labeling_ind}
The goal of this task is to assign a quality score for each individual argument. 
Annotators were presented 
with 
the following binary question per 
argument:\\
\\
\noindent
\setlength{\fboxrule}{1.5pt}
\fbox{
  \parbox{18.60em}{
    Disregarding your own opinion on the topic, would you recommend a friend preparing a speech supporting/contesting the topic to use this argument \emph{as is} in the speech? (yes/no)
  }
}
\\
\\
\noindent All arguments that were collected as described in Section \ref{Data} were labeled in this task. We model the quality of each individual argument as a real value in the range of $[0,1]$, by calculating the fraction of `yes' answers. 
To ensure the annotators will carefully read each argument, the labeling of each argument started with a test question about the stance of the argument towards the concept 
(\textit{pro} or \textit{con}).
The annotators' performance on these test questions was used in the quality control process described in Section \ref{sec:quality}, and also in determining which pairs of arguments to label.

\subsection{Labeling Argument Pairs}
\label{sec:labeling_pairs}
In this task, annotators were presented with a pair of arguments, having the same stance towards the 
concept 
(to reduce bias due to the annotator's opinion), and were asked the following:\\
\\
\setlength{\fboxrule}{1.5pt}
\fbox{
  \parbox{18.6em}{
    Which of the two arguments would have been preferred by most people to support/contest the topic?
  }
}
\\
\\
\noindent Table \ref{table:example} presents an example of such an argument pair
in which the annotators unanimously preferred the first argument.
\begin{table}[h]
\begin{center}
\small
\begin{tabular}{ p{3cm}p{3cm}  }
 \hline
\textbf{Argument 1} & \textbf{Argument 2}\\
\hline
 Children emulate the media they consume and so will be more violent if you don't ban them from violent video games & These are less fun and more harmful games but specifically violent games are played in groups and exclude softer souls \\
 \end{tabular}
 \end{center}
\caption{An example of an argument pair for the motion \textit{We should ban the sale of violent video games to minors}. The first argument was unanimously preferred by all annotators.}
\label{table:example}
\end{table}

As mentioned, annotating all pairs in a large collection of arguments is often not feasible. Thus, we focused our attention on pairs that are presumably most valuable to train a learning algorithm. Specifically, we annotated $14k$ randomly selected pairs, that satisfy the following criteria:
\begin{enumerate}
\item At least $80\%$ of the annotators agreed on the stance of each 
argument, aiming to focus on clearly stated arguments.
\item Individual quality scores in each pair differ by at least $0.2$, aiming for pairs with a 
relatively high 
chance of a clear winner.
\item The length of both arguments, as measured by number of tokens, differs by $\leq 20\%$, 
aiming to focus the task on dimensions beyond argument length. 
\end{enumerate}

\section{Quality Control} 

\label{sec:quality}
To monitor and ensure the quality of collected annotations, we employed the following analyses:\\

\textbf{Kappa Analysis} -- 

\begin{enumerate}
  \item Pairwise Cohen's kappa ($\kappa$) \cite{Cohen1960} is calculated for each pair of annotators that share {\it at least\/} $50$ common argument/argument pairs judgments, and based only on those common judgments.
  \item \textit{Annotator-}$\kappa$ is obtained by averaging all pairwise $\kappa$ for this annotator as calculated in Step 1, 
  and if and only if this annotator had $\geq 5$ pairwise $\kappa$ values estimated. This is used to ignore annotators as described later.
  \item Averaging all \textit{Annotator-}$\kappa$, calculated in Step 2, results in \textit{Task-Average-}$\kappa$.\footnote{It is noted that some annotators remain without valid \textit{Annotator-}$\kappa$ and cannot be filtered out based on their $\kappa$. Similarly, those annotators do not contribute to the \textit{Task-Average-}$\kappa$. However, 
  in both annotation tasks, those annotators contributed only $0.01-0.03$ of the judgments collected.}
  \end{enumerate}


\textbf{Test Questions Analysis} -- Hidden embedded test questions, based on ground truth, are 
often valuable 
for monitoring crowd work. In our setup, at least one fifth of the judgments provided by each annotator are on test questions. When 
annotators fail a test question, they are alerted. Thus, 
beyond monitoring annotator's quality, test questions also provide annotators feedback on task expectations.
In addition, an annotator that fails more than a pre-specified fraction (e.g., $20\%$) 
of the test questions is removed from the task, and his judgments are ignored.

\textbf{High Prior Analysis} -- An annotator that always answers 'yes' to a particular question should obviously be ignored; more generally, we discarded the judgments contributed by annotators with a relatively 
high prior to answer positively on the presented questions. 

Note, if an annotator is discarded due to failure in any of the above analyses, he is further discarded from the estimation of \textit{Annotator-}$\kappa$ and \textit{Task-Average-}$\kappa$.

\section{Data Cleansing} 

\subsection{Cleansing of Individual Arguments Judgments} 
To enhance the quality of the collected data, we discard judgments by annotators who (1) failed $\geq 20\%$ of the test-questions\footnote{Since quality judgments are relatively subjective, we focused the test questions on the stance question.}; and/or (2) obtained \textit{Annotator-}$\kappa \leq 0.35$ in the stance judgment task; and/or (3)  answered `yes' for $\geq 80\%$ of the quality judgment questions. Finally, we discarded arguments that were left with less than $7$ valid judgments. 
This process left us with $5.3k$ arguments, each with $11.4$ valid annotations on average. 
The \textit{Task-Average-}$\kappa$ was $0.69$ on the stance question and $0.1$ on the quality 
question. We refer to the full, unfiltered, set as \textit{IBM-ArgQ-6.3kArgs}, and to the filtered set as \textit{IBM-ArgQ-5.3kArgs} (\textit{IBMRank}).

For completeness, we also attempted to utilize an alternative data cleansing tool, MACE \cite{hovyMACE}. We ran MACE with a threshold k, keeping the top k percent of arguments according to their entropy. We then re-calculated \textit{Task-Average-}$\kappa$ on the resulting dataset. We ran MACE with k=0.95, as used in \newcite{Gurevych16}, and with k=0.85, as this results in a dataset similar in size to \textit{IBMRank}. The resulting \textit{Task-Average-}$\kappa$ is $0.08$ and $0.09$, respectively, lower than our reported $0.1$. We thus maintain our approach to data cleansing as described above. 

The low average $\kappa$ of $0.1$ for quality judgments is expected due to the subjective nature of the task, but nonetheless requires further attention. 
Based on the following observations, we argue that the labels inferred from these annotations are still meaningful and valuable: (1) the high \textit{Task-Average-}$\kappa$ on the stance task conveys the annotators carefully read the arguments before providing their judgments; (2) we report high agreement of the individual quality labels with the argument-pair annotations, to which much better $\kappa$ values were obtained (see Section \ref{annotationSchemes}); (3) we demonstrate that the collected labels can be successfully used by a neural network to predict argument ranking (see section \ref{exp:ranking}), suggesting these labels carry a real signal related to arguments' properties.

\subsection{Cleansing of Argument Pair Labeling}
To enhance the quality of the collected 
pairwise 
data, we discard judgments by annotators who (1) failed $\geq 30\%$ of the test-questions; and/or (2) obtained 
\textit{Annotator-}$\kappa \leq 0.15$ in this task.
Here, the test questions were directly addressing the (relative) quality judgment of pairs, and not the stance of the arguments. In initial 
annotation rounds the test questions were created based on the previously collected individual arguments labels - considering pairs in which the difference in individual quality scores was $\geq 0.6$.\footnote{Note, that annotators have the option of contesting problematic test questions, and thus unfitting ones were disabled during the task by our team.} 
In following annotation rounds, the test questions were defined based on pairs for which $\geq 90\%$ of the annotators agreed on the winning pair. Following this process we were left with an average of $15.9$ valid annotations for each pair, and with \textit{Task-Average-}$\kappa$ of $0.42$ on the quality judgments -- a relatively high value for such a subjective task. As an additional cleansing, for training the learning algorithms, we considered only pairs for which $\geq 70\%$ of the annotators agreed on the winner, leaving us with a total of $9.1k$ pairs for training and evaluation.
We refer to the full, unfiltered, set as \textit{IBM-ArgQ-14kPairs}, and to the filtered set as \textit{IBM-ArgQ-9.1kPairs}.

\section{Data Consistency} 
\subsection{Consistency of Labeling Tasks}
\label{annotationSchemes}
Provided with both individual and pairwise quality labeling, we estimated the consistency of these two approaches. For each pair of arguments, we define the \textit{expected winning argument} as the one with the higher {\it individual\/} argument score, and compare that to the {\it actual winning argument\/}, namely the argument preferred by most annotators when considering the pair directly. Overall, in $75\%$ of the pairs the actual winner was the expected one. Moreover, when focusing on pairs in which the individual argument scores differ by $>0.5$, this agreement reaches $84.3\%$ of  pairs.

 \subsection{Reproducibility Evaluation}
An important property of a valid annotation is its reproducibility. For this purpose, a random sample of $500$ argument pairs from the \textit{IBMPairs} dataset was relabeled by the crowd. This relabeling took place a few months after the main annotation tasks, with the exact task and data cleansing methods that were employed originally. For measuring correlation, 
the following $A\_score$ was defined: the fraction of valid annotations selecting ``argument $A$" in an argument pair $(A,B)$ as having higher quality, out of the total number of valid annotations. Pearson's correlation coefficient between $A\_score$ in initial and secondary annotation of the defined sample was $0.81$.

\begin{figure}[h]

\includegraphics[width=8cm]{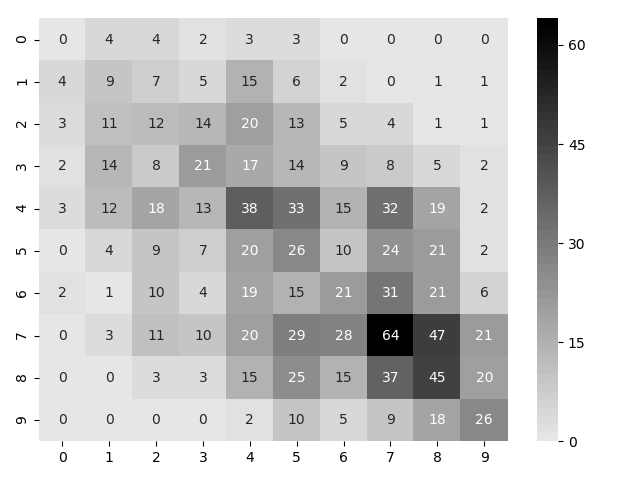}
\caption{Counts of quality score bins between two equally sized sets of annotators.}
\label{heatmap}
\end{figure}

A similar process was followed with the individual arguments quality labeling. Instead of relabeling, we split existing annotations to two even groups. We chose only individual arguments in which at least $14$ valid annotations remained after data cleansing ($1,154$ such arguments). This resulted in two sets of labels for the same data, each based on at least $7$ annotations. Pearson's correlation coefficient between quality scores of the two sets was $0.53$.
We then divided the quality score, which ranges between 0 to 1, to 10 equal bins. The bin frequency counts between the two sets are displayed in the heatmap in Figure \ref{heatmap}.

\subsection{Transitivity Evaluation}
Following \newcite{Gurevych16}, we further examined to what extent our labeled pairs satisfy \emph{transitivity}. Specifically, a triplet of arguments $(A,B,C)$ in which $A$ is preferred over $B$, and $B$ is preferred over $C$, is considered transitive if and only if $A$ is also preferred over $C$. 
We examined all $892$ argument triplets for which all pair-wise combinations were labeled, and found that transitivity holds in $96.2\%$ of the triplets, further strengthening the validity of our data.

\section{Comparison of \textit{IBMRank} and \textit{UKPRank}}
\label{sec:data-comp}
A distinctive feature of our \textit{IBMRank} dataset is that it was collected actively, via a dedicated user interface with clear instructions and enforced length limitations. Correspondingly, we end up with cleaner texts, that are also more homogeneous in terms of length, compared to the \textit{UKPRank} that relies on arguments collected from debate portals.

\subsubsection*{Text Cleanliness}
We counted tokens representing a malformed span of text in \textit{IBMRank} and \textit{UKPRank}. These are HTML markup tags, links, excessive punctuation\footnote{Sequences of three or more punctuation characters, e.g. \textit{``?!?!?!"}}, and tokens not found in GloVE vocabulary \cite{Pennington14glove}. Our findings show that 94.78\% of \textit{IBMRank} arguments contain no malformed text, 4.38\% include one such token, and 0.71\% include two 
such 
tokens. In the case of \textit{UKPRank}, only 62.36\% of the arguments are free of malformed text, 17.59\% include one such token, and 20.05\% include two or more tokens of malformed text.

\subsubsection*{Text Length}
\begin{figure}[htb]
\begin{center}
\includegraphics[height=13em,width=20em]{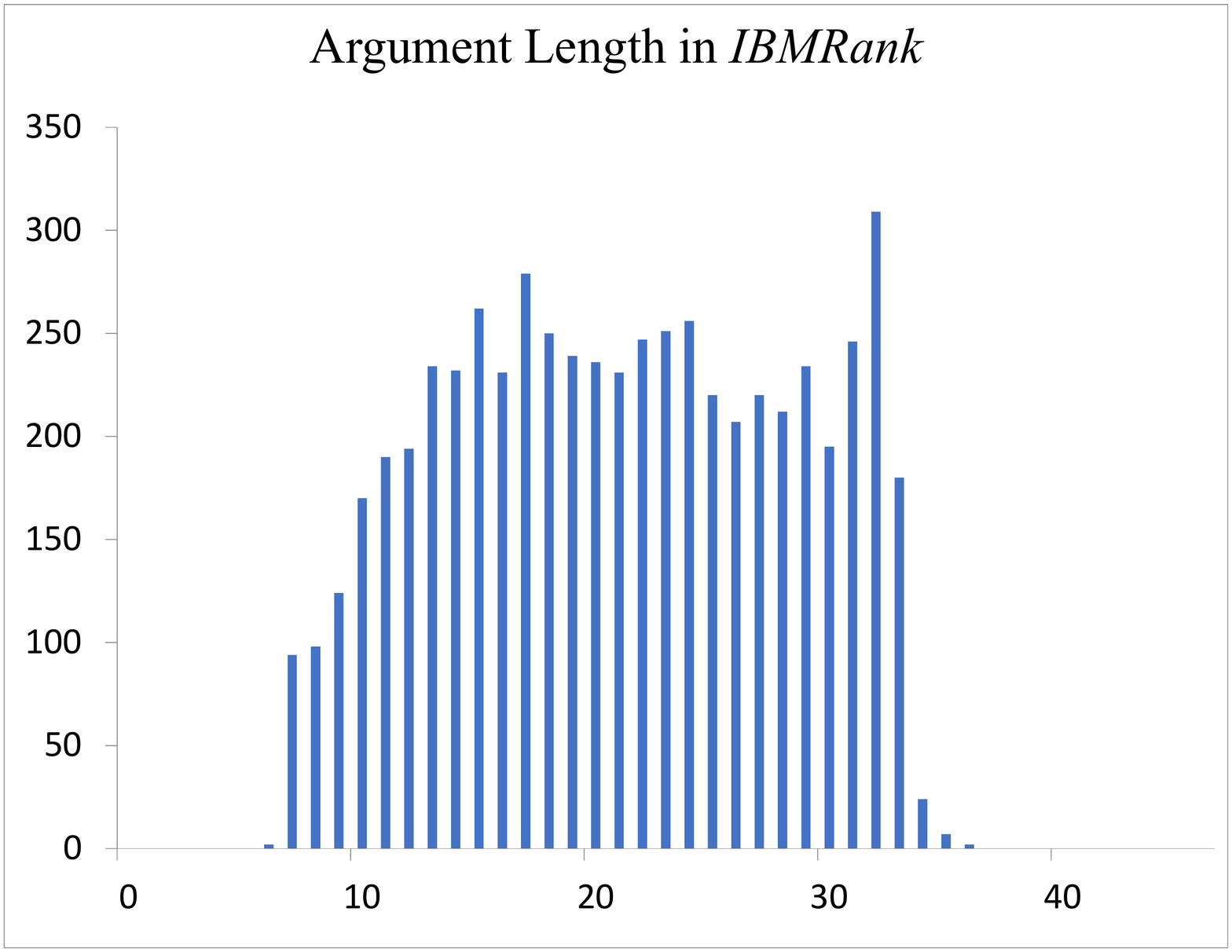}
\includegraphics[height=13em,width=20em]{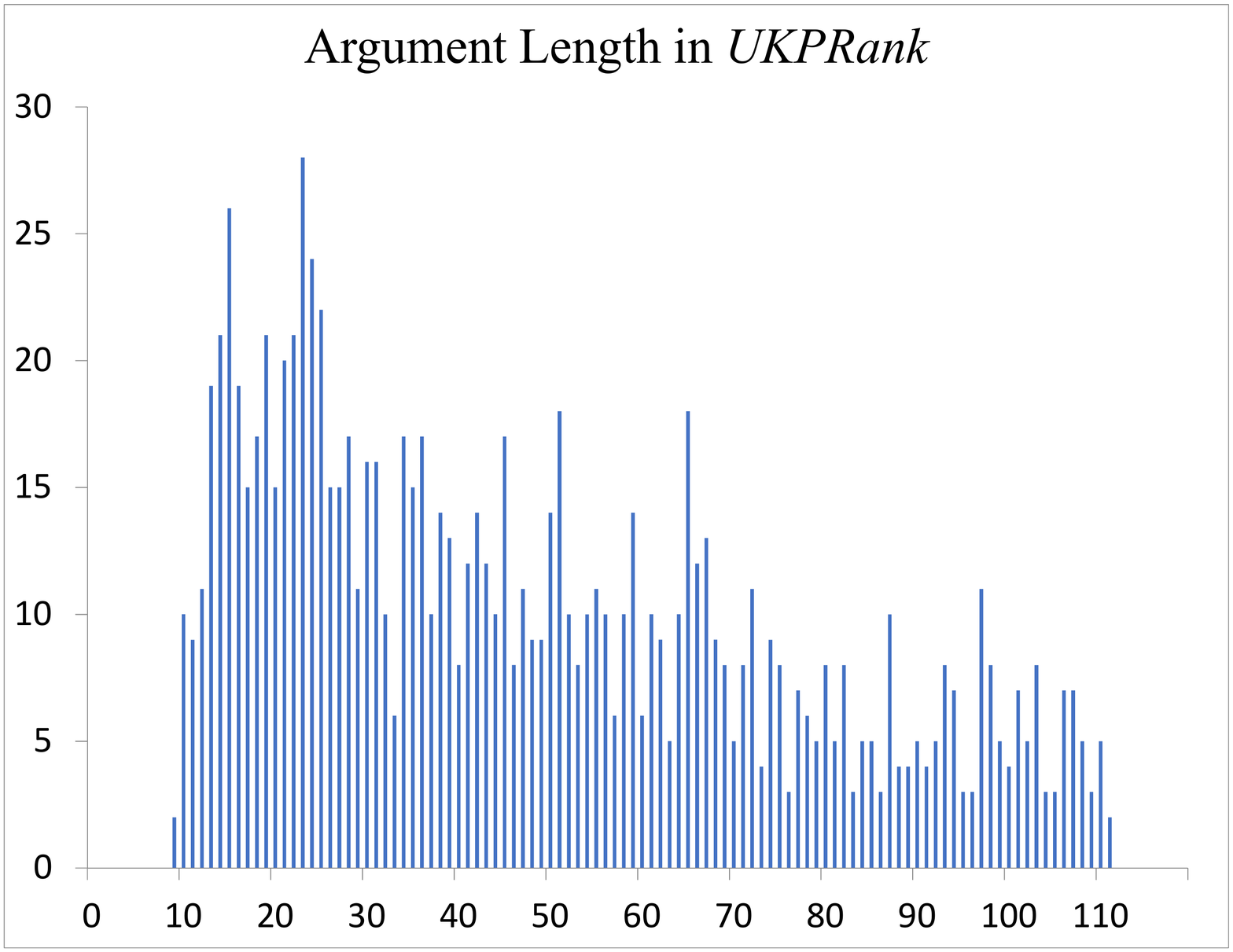}
\caption{Histograms of argument length in 
\textit{IBMRank} and \textit{UKPRank}. X-axis: length (token count). Y-axis: the number of arguments at that length.}
\label{fig:len}
\end{center}
\end{figure}

As depicted in Figure \ref{fig:len}, the arguments in \textit{IBMRank} are substantially more homogeneous in their length compared to \textit{UKPRank}. A potential drawback of the length limitation is that it possibly prevents any learning system from being able to model long arguments correctly. However, by imposing this restriction we expect our quality labeling to be less biased due to argument length, holding greater potential to reveal other properties that contribute to argument quality. We confirmed this intuition with respect to the argument pair labeling as described in Section \ref{sec:argpair}.



\subsubsection*{Data Size and Individual Argument Labeling}
Finally, \textit{IBMRank} covers $5,298$ arguments, compared to $1,052$ in \textit{UKPRank}. In addition, in \textit{UKPRank} no individual labeling is provided, and individual quality scores are inferred from pairs labeling. In contrast, for \textit{IBMRank} each argument is individually labeled for quality, and we explicitly demonstrate the consistency of these individual labeling with the provided pairwise labeling. 

\noindent

\section{Methods}
\label{methods}
In this section we describe neural methods for predicting the individual score and the pair-wise classification of arguments. We devise two methods corresponding to the two newly introduced datasets. Our methods are based upon a powerful language representational model named Bidirectional Encoder Representations from Transformers (BERT) \cite{BERT18} which achieves state-of-the-art results on a wide range of tasks in NLP (\newcite{GLUE}, \newcite{SQuAD11,SQUAD20}). BERT has been extensively trained over large corpora to perform two tasks: (1) \textit{Masked Language Model} - randomly replace words with a predefined token, \verb|[MASK]|, and predict the missing word. (2) \textit{Next Sentence Prediction} - given a pair of sentences \verb|A| and \verb|B|, predict whether sentence \verb|B| follows sentence \verb|A|. Due to its bidirectional nature, BERT achieves remarkable results when fine-tuned to different tasks without the need for specific modifications per task. For further details refer to \newcite{BERT18}.

 


\subsection{Argument-Pair Classification}

We fine-tune BERT's Base Uncased English pre-trained model for a binary classification task.\footnote{Initial experiments with BERT's Large model showed only minor improvements, so for the purpose of the experiments detailed in Section \ref{experiments} we used the Base model.} The fine-tuning process is initialized with weights from the general purpose pre-trained model and a task specific weight matrix $W_{out}\in\mathbb{R}^{768\times 2}$ is added to the $12$-layer base network. Following standard practice with BERT, given a pair of  arguments \verb|A| and \verb|B|, we feed the network with the following sequence `\verb|[CLS]A[SEP]B|'. The \verb|[SEP]| token indicates to the network that the input is to be treated as a pair and \verb|[CLS]| is a token which is used to obtain contextual embedding for the entire sequence.
The network is trained for $3$ epochs with a learning rate of $2^{-5}$. We refer to this model as \textit{Arg-Classifier}.


\subsection{Argument Ranking}
For training a model to output a score between $[0,1]$ we obtain contextual embeddings from the Arg-Classifier fine-tuned model. We concatenate the last 4 layers of the model output to obtain an embedding vector of size $4\times768=3072$.
The embedding vectors are used as input to a neural network with a single output and one hidden layer with $300$ neurons. In order for the network to output values in $[0,1]$, we use a sigmoid activation, $\sigma_{sigmoid}(x)=\frac{1}{1+e^{-x}}$. Denote the weight matrices $W_1\in \mathbb{R}^{3072\times 300}$ and $W_2\in \mathbb{R}^{300\times 1}$, the regressor model, $f_R$, 
is a 2-layered neural network with $\sigma_{relu}(x)=\max\lbrace 0,x \rbrace$ activation.
$f_R$ can be written as:\footnote{We omit bias terms for readability.}
$$f_{R}(x)=\sigma_{sigmoid} \left(W_2^T\sigma_{relu}(W_1^Tx) \right)$$
where $x\in\mathbb{R}^{3072}$ is the embedding vector representing an argument.
We refer to this regression model as \textit{Arg-Ranker}.

\section{Experiments}
\label{experiments}
\subsection{Argument-Pair Classification}
\label{sec:argpair}
In this section we evaluate the methods described in Section \ref{methods}. First, we evaluate the accuracy of \textit{Arg-Classifier} on our \textit{IBMPairs} dataset and on \textit{UKPConvArgStrict} (henceforth, \textit{UKPStrict}), the filtered argument pairs dataset of \newcite{Gurevych16}, in k-fold cross-validation.\footnote{$22$ and $32$ folds respectively.}
We calculate accuracy and ROC area under curve (AUC) for each fold, and report the weighted averages over all folds. We also evaluate \newcite{Gurevych18}'s \textit{GPPL} median heuristic method with \textit{GloVe} + \textit{ling} features in cross-validation on our \textit{IBMPairs} dataset. For completeness, we quote \newcite{Gurevych18}'s figures of \textit{GPPL opt.} and \textit{GPC} on \textit{UKPStrict}.\footnote{We were unable to reproduce the results reported in \newcite{Gurevych18} by running the \textit{GPPL opt.} and \textit{GPC} algorithms on the \textit{UKPStrict} dataset. We have approached the authors and reported the issue, which was not solved by the time this paper was published, and hence we only quote the figures as reported there.} 
We add a simple baseline classifying arguments based on their token count (\textit{Arg-Length}). 

\begin{table}[h]
\begin{center}
\scriptsize
\begin{tabular}{ |c|c|c|c|  }
\hline
\multirow{2}{*}{} & \multicolumn{3}{c|}{IBMPairs}\\ \cline{2-4} 
 & Arg-Length & Arg-Classifier & GPPL \\
 \hline
  Acc. & $.55$ & $\textbf{.80}$ & $.71$ \\
 \hline
 AUC & $.59$ & $\textbf{.86}$ & $.78$ \\
 \hline
 \end{tabular}
 
\bigskip

\begin{tabular}{|c|c|c|c|c|c|}
\hline
\multirow{2}{*}{} & \multicolumn{5}{c|}{UKPStrict}\\ \cline{2-6}
& Arg-Length & Arg-Classifier & GPPL & GPPL opt. & GPC \\
 \hline
 Acc. & $.76$ & $\textbf{.83}$ & $.79$ & $.80$ & $.81$\\
 \hline
AUC & $.78$ & $\textbf{.89}$ & $.87$ & $.87$ & $\textbf{.89}$\\
 \hline
 \end{tabular}
 
 \end{center}
 \caption{Accuracy and AUC on \small{\textit{IBMPairs}} and \small{\textit{UKPStrict}}.}
\label{table:pairs}
\end{table}

As can be seen in Tables \ref{table:pairs}, \textit{Arg-Classifier} improves on the \textit{GPPL} method on both datasets ($p\ll.01$ using two-tailed Wilcoxon signed-rank test).\footnote{\label{perfold}The results per fold in both tasks are included in the supplementary material.} We note that \textit{Arg-Classifier}'s accuracy on the \textit{UKPStrict} set is higher than all methods tested on this dataset in \newcite{Gurevych16,Gurevych18}. Interestingly, all methods reach higher accuracy on \textit{UKPStrict} compared to \textit{IBMPairs}, presumably indicating that the data in \textit{IBMPairs} is more challenging to classify. With regards to \textit{Arg-Length}, we can see that it is inaccurate on \textit{IBMPairs} but achieves a respectable result on \textit{UKPStrict}. This is in agreement with  \newcite{habernal-gurevych-2016-makes} who analyzed the reasons that annotators provided for their labeling. In most cases the reason indicated preference for arguments with more information -- which is what longer arguments tend to be better at. This further strengthens the value of creating \textit{IBMPairs} and \textit{IBMRank} as much more homogeneous datasets in terms of argument length.

\subsection{Argument Ranking}
\label{exp:ranking}
We proceed to evaluate the \textit{Arg-Ranker} on the \textit{IBMRank} and \textit{UKPRank} datasets in k-fold cross-validation, and report weighted correlation measures. We also evaluate the \textit{Arg-Ranker} by feeding it vanilla BERT embeddings, instead of the fine-tuned embeddings generated by the \textit{Arg-Classifier} model.  We refer to this version as \textit{Arg-Ranker-base}. In both \textit{Arg-Ranker} and \textit{Arg-Ranker-base} evaluations we report the mean of 3 runs.\footnote{The \textit{GPPL} regressor of \newcite{Gurevych18} relies on pair-wise (relative) labeling of arguments and as a result it cannot be used for predicting the  individual (absolute) labeling of arguments, as in \textit{IBMRank}.}

\begin{table}[h]
\begin{center}
\scriptsize
\setlength\tabcolsep{0.08cm}
\begin{tabular}{ |C{0.1cm}|C{1.7cm}|C{1.20cm}|C{1.7cm}|C{1.20cm}|C{0.58cm}|  }
 \hline
 \multirow{2}{*}{} & \multicolumn{2}{c|}{IBMRank} & \multicolumn{3}{c|}{UKPRank} \\
 \cline{2-6}
 & Arg-Ranker-base & Arg-Ranker & Arg-Ranker-base & Arg-Ranker & GPPL \\
 \hline
 $r$ & $.41$ & $\textbf{.42}$ & $.44$ & $\textbf{.49}$ & $.45$\\
 \hline
 $\rho$ & $.38$ & $\textbf{.41}$ & $.57$  & $.59$ & $\textbf{.65}$\\
 \hline
 \end{tabular}
 \end{center}
 \caption{Pearson's ($r$) and Spearman's ($\rho$) correlation of \textit{Arg-Ranker-base}, \textit{Arg-Ranker} and \textit{GPPL} on the \textit{IBMRank} and \textit{UKPRank} datasets.}
\label{table:inv}
\end{table}

As can be seen in Table \ref{table:inv}, on the \textit{UKPRank} dataset, \textit{Arg-Ranker} is slightly better than \textit{GPPL} for Pearson's correlation, but slightly worse for Spearman's correlation. Additionally,  using direct BERT embeddings provides worse correlation\footnote{Significantly for the \textit{IBMRank} data on both measures, and for the \textit{UKPRank} on Pearson's correlation, $p\ll.05$.} than using the \textit{Arg-Classifier} embeddings for both datasets, justifying its use. Finally, similarly to the findings in the argument-pair classification task, the \textit{IBMRank} dataset is harder to predict.\footnote{For the experiments on \textit{IBMRank}, we included by mistake a small fraction of arguments which actually should have been filtered. The effect on the results is minimal.}

\section{Error Analysis}

We present a qualitative analysis of examples that the \textit{Arg-Classifier} and \textit{Arg-Ranker} models did not predict correctly. For each of the argument-pair and ranking tasks, we analyzed $50-100$ arguments from three motions on which the performance of the respective model was poor. For each motion we selected the arguments in which the model was most confident in the wrong direction.

\begin{table*}[h]
\begin{center}
\small
\begin{tabular}{ |p{2cm}|p{1.5cm}|p{5cm}|p{5cm}|  }
 \hline
Motion & Type & Argument1 & Argument2\\ \cline{1-3}
 \hline
We should ban fossil fuels & Impact over delivery &\textit{the only way to provide any space for energy alternatives to enter the market is by artificially decreasing the power of fossil fuels through a ban.} & \textbf{fossil fuels are bad for the environment, they have so2 in them that is the thing that maks acid rain and it is today harming the environment and will only be wors.}  \\
 \hline
Flu vaccination should not be mandatory & Provocative or not grounded & \textbf{the only responsible persons for kids are their parents. if they dont think that their kids should get the vaccine its their own decision.} & \textit{the body has an automatic vaccination due to evolution, those who got sick and died are the weakest link and we are better off without them} \\
\hline
We should abandon vegetarianism & Consistent annotator preference & \textbf{it's harder to get all the things you need for a balanced diet while being vegetarian.} & \textit{animals deserve less rights than humans, and it is legitimate for humans to prioritize their enjoyment over the suffering of animals.} \\
\hline
 \end{tabular}
 \end{center}
 \caption{Examples of argument pairs for which there is a high difference between the argument selected by the annotators, marked in bold, and the argument predicted to be of higher quality by the model, marked in italics.}
\label{table:errorAnalysis}
\end{table*}

A prominent insight from this analysis, common to both models, is that the model tends to fail when the argument persuasiveness outweighs its delivery quality (such as bad phrasing or typos). 
An example of this is shown in row 1 of Table \ref{table:errorAnalysis}. In this case, Argument2 is labeled as having a higher quality, even though it contains multiple typos, and thus is typical to arguments that the model was trained to avoid selecting.

Another phenomenon that both our models fail to address is arguments that are off-topic, too provocative or not grounded. An example of this, from the argument-pair task, is shown in row 2 - Argument2 is presumably considered harsh by annotators, even though it is fine in terms of grammatical structure and impact on the topic. These types of arguments are becoming more important to recognize, especially in the ``fake-news" era. We leave dealing with them for future work.

Finally, we also notice certain arguments were consistently preferred by annotators, regardless of the quality of the opposing argument. This is a pattern relevant only to the \textit{Arg-Classifier} model, shown in row 3.


\section{Conclusions and Future Work}

A significant barrier in developing automatic methods for estimating argument quality is the lack of suitable data. 
An important contribution of this work is a newly introduced data composed of $6.3k$ carefully annotated
arguments, compared to $1k$ arguments in previously considered data. 
Another barrier is the inherent subjectivity of the {\it manual\/} task for determining argument quality. 
To overcome this issue, we employed a relatively large set of crowd annotators to consider each instance, associated
with various measures to ensure the quality of the annotations associated with the released data. 
In addition, while previous work focused on arguments collected from web debate portals, here we collected
arguments via a dedicated interface, enforcing length limitations, and providing contributors with clear guidance.
Moreover, previous work relied solely on annotating pairs of arguments, and used these annotations to {\it infer\/} the individual ranking of arguments; in contrast, here, we annotated all individual arguments for their quality, and further annotated $14k$ pairs. This two--fold approach allowed us, for the first time, to explicitly examine the relation between relative (pairwise) annotation and explicit (individual) annotation of argument quality. Our analysis suggests that these two schemes provide relatively 
consistent results. In addition, these annotation efforts may complement each other. As pairs of arguments with a high difference in individual quality scores appear to agree with argument-pair annotations, one may deduce the latter from the former. Thus, it may be beneficial to dedicate the more expensive pair-wise annotation efforts to pairs in which the difference in individual quality scores is small, reminiscent of active learning \cite{settles2009active}. In future work we intend to further investigate this approach, as well as explore in more detail the low fraction of cases where these two schemes led to clearly different results.

The second contribution of this work is suggesting neural methods, based on \newcite{BERT18}, for argument ranking as well as for argument-pair classification. In the former task, our results are comparable to state-of-the-art; in the latter task they significantly outperform earlier methods \cite{Gurevych16}.

Finally, to the best of our knowledge, current approaches do not deal with argument pairs of relatively {\it similar\/} quality. A natural extension is to develop a ternary-class classification model that will be trained and evaluated on such pairs, as we intend to explore in future work.

\section*{Acknowledgements}
We thank Tel Aviv University Debating Society, Ben Gurion University Debating Society, Yale Debate Association, HWS Debate Team, Seawolf Debate Program of the University of Alaska, and many other individual debaters.

\bibliography{argquality_emnlp2019}
\bibliographystyle{acl_natbib}

\end{document}